\documentclass[10pt, a4paper]{article}

\usepackage[]{lrec-coling2024} 

\usepackage{graphicx}
\usepackage{algorithm}
\usepackage{algorithmic}
\usepackage{xcolor}
\usepackage{soul}
\usepackage{makecell}
\usepackage{booktabs}
\usepackage{color}

\title{SDA: Simple Discrete Augmentation for Contrastive Sentence Representation Learning}

\name{Dongsheng Zhu$^{1,3,\textasteriskcentered}$, Zhenyu Mao$^{2,\textasteriskcentered}$, Jinghui Lu$^{2,}$\sthanks{\ \ Equal contribution.}, Rui Zhao$^{2}$, Fei Tan$^{2,}$\sthanks{\ \ Corresponding author.}} 

\address{
    $^{1}$ Baidu Inc. 
    $^{2}$ SenseTime Research \\
    $^{3}$ Fudan University \\
    \texttt{dszhu20@fudan.edu.cn} \\
    \texttt{\{maozhenyu, lujinghui1, zhaorui, tanfei\}@sensetime.com}
    }

\abstract{
Contrastive learning has recently achieved compelling performance in unsupervised sentence representation. As an essential element, data augmentation protocols, however, have not been well explored. The pioneering work SimCSE resorting to a simple dropout mechanism (viewed as \textit{\textbf{continuous}} augmentation) surprisingly dominates discrete augmentations such as cropping, word deletion, and synonym replacement as reported. To understand the underlying rationales, we revisit existing approaches and attempt to hypothesize the desiderata of reasonable data augmentation methods: balance of semantic consistency and expression diversity. We then develop three simple yet effective \textit{\textbf{discrete}} sentence augmentation schemes: \textit{\textbf{punctuation insertion}}, \textit{\textbf{modal verbs}}, and \textit{\textbf{double negation}}. They act as minimal noises at lexical level to produce diverse forms of sentences. Furthermore, \textit{\textbf{standard negation}} is capitalized on to generate negative samples for alleviating feature suppression involved in contrastive learning. We experimented extensively with semantic textual similarity on diverse datasets. The results support the superiority of the proposed methods consistently. Our key code is available at \href{https://github.com/Zhudongsheng75/SDA}{https://github.com/Zhudongsheng75/SDA}
 \\ \newline  }

\begin{document}

\maketitleabstract

\section{Introduction}

Current state-of-the-art methods utilize contrastive learning algorithms to learn unsupervised sentence representations \cite{gao2021simcse,yan2021consert}. They learn to bring similar sentences closer in the latent space while pushing away dissimilar ones \cite{hjelm2018learning}. In the paradigm, multiple data augmentation methods have been proposed from different perspectives to curate different variants\footnote{It refers to a sentence sample in this work.} \cite{oord2018representation,zhu2020deep}. The variants and the corresponding original samples are deemed positive pairs in the learning procedure \cite{chen2020simple}. Previous studies have shown that the quality of variants largely shapes the learning of reasonable representations \cite{chen2020simple,hassani2020contrastive}. 

\begin{figure}[t]
\centering
\includegraphics[width=1\columnwidth]{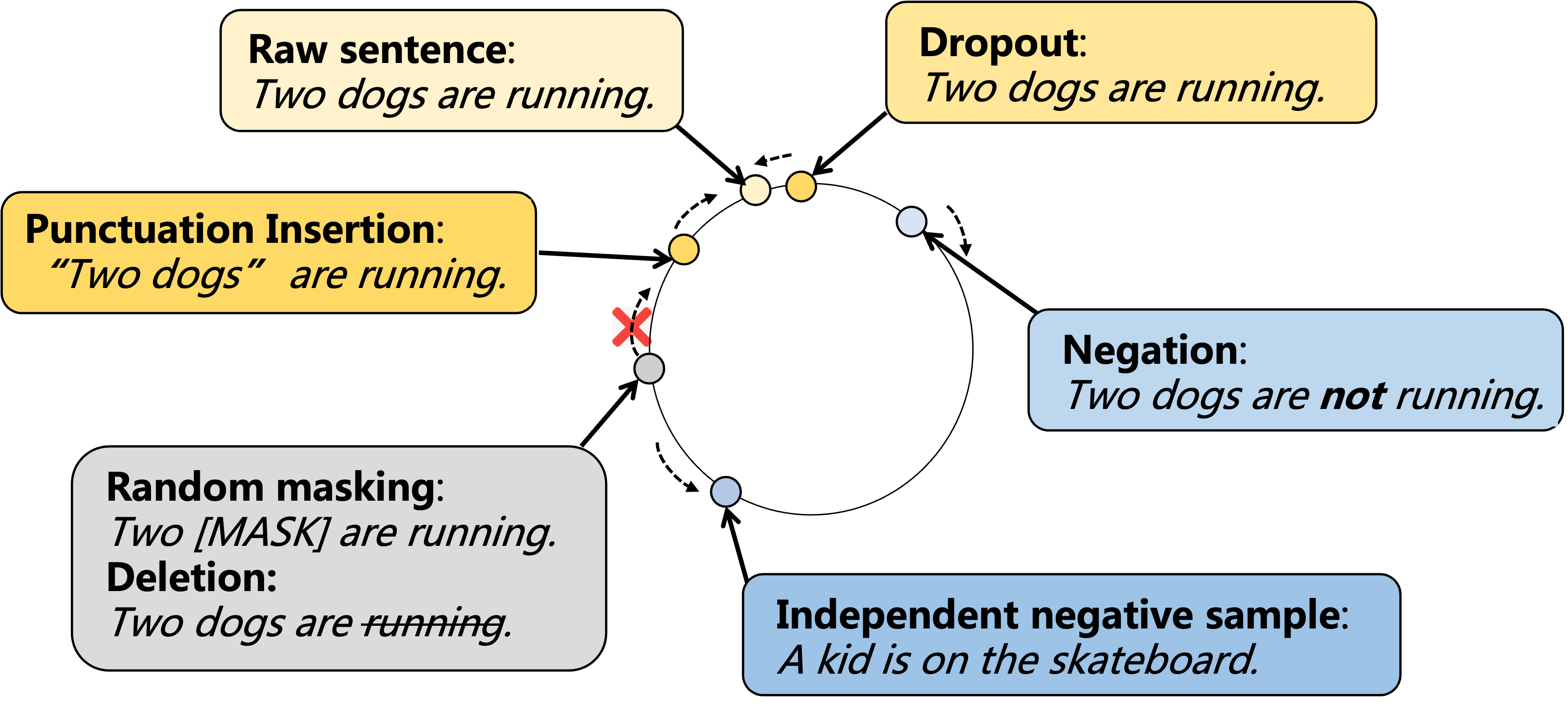}
\caption{Normalized representation visualization of different augmentation methods and the way they should be optimized.}
\label{example}
\end{figure}

Conventional methods directly employ operations such as cropping, word deletion, and synonym replacement in natural sentences \cite{wei2019eda,wu2020clear,meng2021coco}. In addition, recent studies resort to network architectures for manipulating embedding vectors, such as dropout, feature cutoff, and token shuffling \cite{gao2021simcse,yan2021consert}. It enables more subtle variants of training samples in the continuous latent space in a controllable way and thereby renders better representations, which is usually evidenced by more appealing performance in typical downstream NLP tasks (e.g., textual semantic similarity) \cite{zhang2021pairwise,chen2021hiddencut}. 

If two sentences with large semantic gaps are paired positively, the representation alignment is likely to deteriorate \cite{wang2020understanding}. Discrete augmentation methods usually have less competitive results, because they can hardly keep the sentence semantically consistent when applied randomly at lexical level. As shown in Figure \ref{example}, sentence semantics become incomplete after random masking and word deletion \cite{wu2020clear} so that optimizing their embeddings in the direction of the original sentence (aka anchor) leads to misunderstanding.

To learn better representations, augmentation methods are supposed to generate samples that are not only diverse in representation \cite{tian2020makes} but also similar in semantics. There is, however, a trade-off between semantic consistency and expression diversity. Bigger differences between vanilla samples and augmented ones also convey less faithful semantics. Therefore, we hypothesize that a good augmentation method in contrastive learning should have desiderata to balance them.

Continuous methods can control semantic changes since they utilize designed network structures to process redundant features \cite{huang2021ghostbert}. SimCSE \cite{gao2021simcse} utilizes dropout \cite{srivastava2014dropout} to obtain different embeddings of the same sentence to construct positive pairs. But such continuous methods lack interpretability to inspire further exploration of sentence augmentation. To better prove our hypotheses and find the promising direction of lexical data augmentation methods, we propose three \textbf{S}imple \textbf{D}iscrete \textbf{A}ugmentation (SDA) methods to satisfy the desiderata to different extents: \textit{Punctuation Insertion} (PI), \textit{Modal Verbs} (MV), and \textit{Double Negation} (DN). Their impacts on the expression diversity increase in a row but semantic consistency with the original sentence tends to diminish gradually. In linguistics, punctuation usually represents pause or tone (e.g., comma, exclamations) which has no specific meaning itself. Modal verbs are used as supplementary to the predicate verb of the sentence indicating attitudes such as permission, request, and so on, which helps to reduce uncertainty in semantics. DN helps two negatives cancel each other out and thereby produces a strong affirmation, whereas the improperly augmented sentence is at risk to be logically confusing. 

Although the proposed augmentation methods keep the semantic meaning by carefully adding minor noises, the generated sentences are still literally similar to the original sentence. Recent research \cite{robinson2021can} has pointed out the feature suppression problem of contrastive learning. This phenomenon could result in shortcut solutions that the model only learns the textual rather than semantic similarity. A focus on hard examples has been proven effective to change the scope of the captured features \cite{robinson2021can}. Thus, we further utilize standard negation to construct text contradicting all or part of the meaning of the original sentence as hard negative samples. By doing so, the model is encouraged to learn to differentiate sentences bearing similar lexical items yet reversed meanings.

To summarize, the contributions of this work are as follows:
\begin{itemize}
    \item We propose SDA methods (including standard negation) for contrastive sentence representation learning, which leverages discrete sentence modifications to enhance the performance of representation learning (Section \ref{sec:method}). 
    \item Comprehensive experimental results demonstrate that SDA achieves significantly better performance, advancing the state-of-the-art performance to a new bar from 78.49 to 79.60 (Section \ref{sec:exp_setting}).
    \item Extensive ablations and in-depth analysis are conducted to investigate the underlying rationale and clarify the hyper-parameters choices (Section \ref{sec:results}).
\end{itemize}

\begin{figure*}[ht]
\centering
\includegraphics[width=0.8\textwidth]{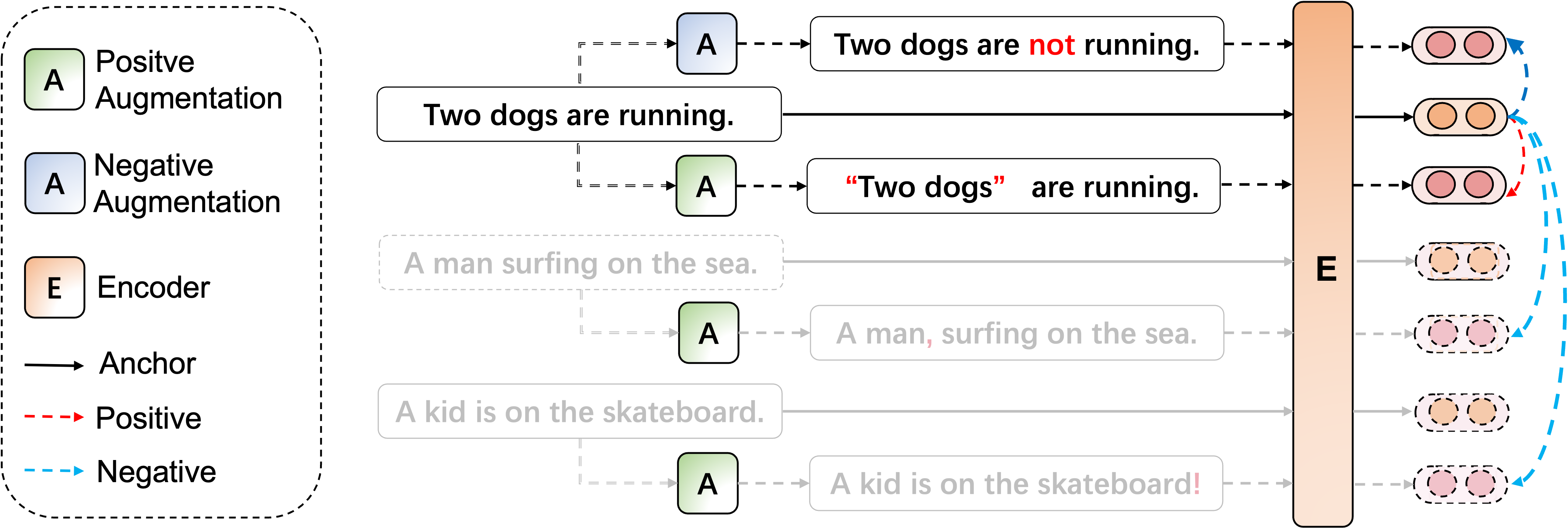}
\caption{An overview of the framework. The figure can be embodied as a training batch. Each sentence is passed through the augmentation module to generate one positive and one negative for the anchor, the positives generated by other sentences in the batch are also deemed as negatives for the anchor.}
\label{overview}
\end{figure*}

\section{Related Works}

\subsection{Sentence Representation Learning}

BERT \cite{devlin-etal-2019-bert} has steered the trajectory of sentence representation towards the technical orientation of Pre-trained Language Models (PLM). A multitude of endeavors \cite{tan2020tnt,tan2021bert,li2020sentence,su2021whitening, lu2023punifiedner, lu2023makes} has been dedicated to substantial improvements based on this paradigm, leading to significant advancements in diverse domains. Notably, there is a pronounced practical demand for sentence-level text representations \cite{conneau2017supervised,williams2018broad}. Consequently, learning unsupervised sentence representations based on PLM has become a focal point in recent years \cite{2019Sentence,zhang2020unsupervised}. Current state-of-the-art methods utilize contrastive learning to learn sentence embeddings \cite{kim2021self,yan2021consert,gao2021simcse}, which in experimental results, can even rival supervised methods. However, to advance unsupervised contrastive learning methods further, data augmentation emerges as a pivotal component.

\subsection{Data Augmentation in Contrastive Learning}

Early research on contrastive sentence representation learning \cite{zhang2020unsupervised} didn't utilize explicit augmentation methods to generate positive pairs. Later, methods \cite{giorgi2021declutr,wu2020clear,wu2021esimcse} which use text augmentation methods, such as word deletion, span deletion, reordering, synonym substitution, and word repetition, to generate different views for each sentence achieve better results. Compared to augmentation methods applied on text, several studies \cite{janson2021semantic,yan2021consert,gao2021simcse,wang2022contrastive} utilize neural networks, such as dual encoders, adversarial attack, token shuffling, cutoff and dropout, to obtain different embeddings for contrasting. A more recent study DiffCSE \cite{chuang2022diffcse} designed an extra MLM-based word replacement detection task as an equivalent augmentation. The purpose of data augmentation in this study is to generate both semantic similar and expression diverse samples so that models can learn to extend the semantic space of the input samples.


\section{Preliminaries}

\subsection{Sentence-level Contrastive Learning}
Given a set of sentence pairs $\mathcal{D}=\{(x_i, x_i^+)\}$, where sentence pairs $(x_i, x_i^+)$ are semantically similar and deemed as positive pairs. Contrastive learning aims to learn a dense representation $\mathbf{h}_i$ of a sentence $x_i$ by gathering positive samples together while pushing others apart in the latent space \cite{belghazi2018mutual}. In practice, the training proceeds within a mini-batch of $N$ sentence pairs. The objective is formulated as:
\begin{equation}
    l_i = -\frac{e^{\mathrm{sim}(\mathbf{h}_i, \mathbf{h}_i^+)/\tau}}{\sum_{j=1}^N e^{\mathrm{sim}(\mathbf{h}_i, \mathbf{h}_j^+)/\tau}}
\label{eq1}
\end{equation}
where $\mathbf{h}_i$ and $\mathbf{h}_i^+$ respectively denote the representation of $x_i$ and $x_i^+$, $\mathrm{sim}()$ is the cosine similarity function and $\tau$ is the temperature parameter. Under the unsupervised setting, the semantically related positive pairs are not explicitly given. Augmentation methods are used to generate $x_i^+$ for training sample $x_i$.

\subsection{Unsupervised SimCSE}
In transformer model $f(\cdot)$, there are dropout masks placed on fully-connected layers and attention probabilities. SimCSE\footnote{The SimCSE mentioned in this article are all under the unsupervised setting.} builds the positive pairs by feeding the same input $x_i$ to the encoder twice, i.e., $x_i^+=x_i$. With different dropout masks $z_i$ and $z_i^+$, the two separate output sentence embeddings constitute a positive pair as follows:

\begin{equation}
    \mathbf{h}_i  =f_{z_i}(x_i), \mathbf{h}_i^+ =f_{z_i^+}(x_i)
\label{eq2}
\end{equation}

\subsection{Dependency Parsing and Syntax Tree}

Dependency parsing represents the relationships between words in a sentence in the form of dependencies. Each word in the sentence is connected to another word, indicating its grammatical role and the type of relationship it has with other words.

Syntax trees represent the hierarchical structure of a sentence's grammar. They consist of nodes, where each node represents a word or a grammatical unit, and edges represent syntactic relationships. The root node represents the main clause, and branches indicate phrases and sub-clauses.

\begin{table}[t]
\small
\centering
\begin{tabular}{ll}
\toprule
  Method & Sentence \\ \hline
  None & He travelled widely in Europe.\\
  PI & He {\color{red}\textbf{,}} travelled widely in Europe.\\
  MV & He {\color{red}\textbf{must have}} travelled widely in Europe.\\
  DN & \makecell[l]{{\color{red}\textbf{It is not the fact that}} he \\
  {\color{red}\textbf{didn't}} travel widely in Europe.}\\
  Negation &he {\color{red}\textbf{didn't}} travel widely in Europe.\\
  \bottomrule
\end{tabular}
\caption{An example of different methods to generate the augmented sentence. The highlighted red texts denote changes after augmentation.}
\label{example}
\end{table}

\section{Methodology} \label{sec:method}

In this work, the augmentation module to generate positive samples for training data is denoted as $A(\cdot)$. As illustrated in Fig. \ref{overview}, we utilize $A(\cdot)$ to subtly reword the original sentence in an attempt to change the representation of the sentence to a limited extent on the premise that the sentence roughly remains unchanged semantically. Afterword, Eq. \ref{eq2} can be rewritten as follows:
\begin{equation}
    \mathbf{h}_i=f_{z_i}(x_i),\mathbf{h}_i^+=f_{z_i^+}(A(x_i))
\label{eq3}
\end{equation}

\begin{figure}
    \centering
    \includegraphics[width=0.8\columnwidth]{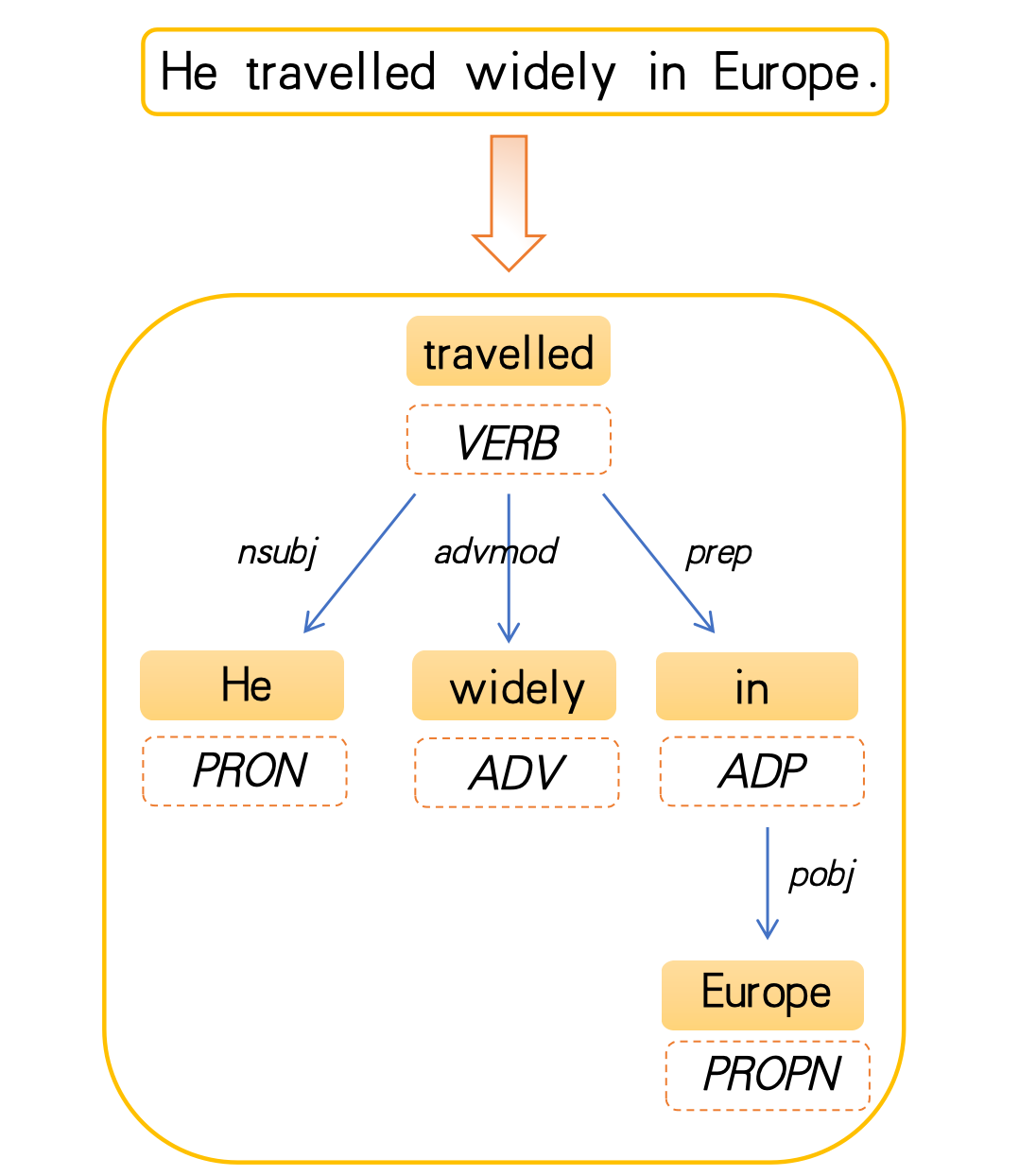}
    \caption{The syntax tree constructed through dependency parsing and its representation.}
    \label{dependency}
\end{figure}

In practice, we utilize spaCy\footnote{https://spacy.io/} for dependency parsing\footnote{We empirically think that grammatically correct sentence transformation is more reasonable than the stochastic manipulation without complying with basic grammar rules explicitly.} and then implement punctuation insertion, modal verbs, and double negation. As shown in Figure \ref{dependency}, first, each sentence needs to be processed through dependency parsing to obtain a central word and the dependencies between the other words with the central word. Then, with the central word as the root node and other words as child nodes, the corresponding syntax tree can be constructed through dependencies. Based on this, we can design rules by traversing the syntax tree in level to implement the SDA augmentation for sentences as shown in Table \ref{example} and Appendix \ref{sec:algo}.

\subsection{Punctuation Insertion (PI)}
Punctuation usually doesn't carry specific information, whereas changing punctuation can have a slight effect on the sentence. We devise a few scalable rules and thereby generate semantically similar examples as positives in a relatively controllable way: (1) If a sentence has a subordinate clause, insert a comma <,> between them; (2) If there is a noun subject in the sentence, enclose it with double quotes <" "> or append a comma <,>; (3) Replacing the punctuation at the end of a sentence with an exclamation point <!> or appending it to the sentence.

These simple rules insert punctuation marks at different positions of a sentence to ensure the diversity of augmentations. Moreover, the rules are context-aware schemes. For example, <!> at the end of sentences expresses emphasis tone, <,> between clauses means to pause, and <" "> frames the subject to highlight it. In this way, PI preserves sentence structure and meaning to the greatest extent and minimizes the magnitude of semantic changes.

\subsection{Modal Verbs (MV)}
Adding modal verbs to a sentence is also a positive generation method that slightly adjusts the meaning of the sentence. The semantic change brought by MV may be greater than punctuation, but it is still linguistically reasonable. 
We first collect a list of modal verbs such as: "must", "should", and "ought to", as well as phrases that have the same function such as "have to", "can't but", "can't help to", etc. Then we randomly pick one phrase to modify the predicate verb of the main clause. MV is more radical than PI because it supplements the speaker's attitudes (e.g., likelihood, permission, obligation, and so on) towards the predicate verbs of the sentence, uncertainty in semantics is thus reduced to some extent.

\subsection{Double Negation (DN)} \label{sec:dn}
In English and Chinese, double negatives cancel one another and produce an affirmative. We leverage this linguistic rule to generate double negation as the positive of the raw sentence. But double negative is difficult to implement and multiple negative words in one sentence could make it confusing. Therefore, DN is considered to be the most radical among all three augmentation methods.

The implementation is to apply a negative transformation to a sentence twice. Either inserting a negative word or removing a negative is regarded as a negative transformation. The negative words are used to modify nouns or verbs in the sentence, and corresponding negative words will be selected, e.g., "no" is inserted before nouns and "not" before verbs. If the sentence is only able to apply negative transformation once, we simply add a negation phrase, such as "It is not true that" or "It can't be that", in front of the sentence as a second negative transformation. 

\subsection{Negation as Negative Sample}
The aforementioned augmentation methods are to generate semantically consistent positive samples for the anchor sentence. To ensure the model can learn the semantic correlation rather than the textual shortcut solutions, we generate a negation sentence for every training sample serving as a negative sample. The negation sentence and the anchor are highly similar in the text but have opposite semantic meanings. Learning to distinguish a negation sentence from the anchor is helpful to alleviate feature suppression \cite{wang2022sncse}. However, since negative samples in contrastive learning are usually independent sentences that are completely irrelevant to the original sentence, it is inappropriate to treat negation sentences the same as those independent negative samples. Therefore, we relax the restriction of the negation sentence in the optimization with a hyper-parameter $\delta$, and rewrite the contrastive loss function in Eq. \ref{eq1} as follows:
\begin{equation}
    l_i = -\frac{e^{\mathrm{sim}(\mathbf{h}_i, \mathbf{h}_i^+)/\tau}}{\sum_{j=1}^N e^{\mathrm{sim}(\mathbf{h}_i, \mathbf{h}_j^+)/\tau} + e^{[\mathrm{sim}(\mathbf{h}_i, \mathbf{h}_i^-)-\delta]/\tau}}
\label{eq4}
\end{equation}
where $\mathbf{h}_i^-$ is the representation of the negation sentence $x_i^-$. And $\delta$ can be considered as a margin which relaxed the distance between sentence $x_i$ and $x_i^-$ compared to other independent negative samples. The implementation of negation transformation has been introduced in Section \ref{sec:dn}, and the rules guarantee the transformation is applicable for arbitrary sentences.

\section{Experiment Settings} \label{sec:exp_setting}


\subsection{Datasets}
Semantic textual similarity (STS) task measures the semantic similarity of any two sentences, it uses Spearman's correlation \cite{myers2013research} to measure the correlation between the ranks of predicted similarities and the ground-truth labels. We conduct our experiments on seven standard English and one Chinese STS datasets. 
For English STS, we follow SimCSE\footnote{https://github.com/princeton-nlp/SimCSE} \cite{gao2021simcse} to use one million sentences randomly drawn from English Wikipedia for training. Datasets STS12-STS16 have neither training nor development sets, and thus we evaluate the models on the development set of STS-B to search for optimal settings of hyper-parameters. For Chinese STS, we choose the unlabeled corpus data in SimCLUE dataset\footnote{https://github.com/CLUEbenchmark/SimCLUE} for training, which contains about 2.2 million sentences. In the test phase, we use the Chinese STS-B from CNSD\footnote{https://github.com/pluto-junzeng/CNSD} without the training set.

\begin{table}[t]
\centering
\begin{tabular}{lc}
\toprule
\textbf{Method} &\textbf{Augmented \%}\\ \hline
Punctuation Insertion   &98.14\% \\
Modal Verbs             &88.32\% \\
Double Negation         &87.89\% \\
\bottomrule
\end{tabular}
\caption{The ratio of three augmentation protocols.}
\label{aug_per}
\end{table}

\begin{table}[t]
\centering
\resizebox{\columnwidth}{!}{
\begin{tabular}{lccc}
\toprule
\textbf{Param} & \textbf{BERT$_{base}$} & \textbf{RoBERTa$_{base}$} & \textbf{CN-RoBERTa$_{base}$} \\ \hline
learning rate & 2e-5 & 1e-5 & 5e-5 \\
batch size & 128 & 64 & 128   \\
$\tau$ & 0.05 & 0.05 & 0.05  \\
$\delta$ & 0.5 & 0.5 & 0.5  \\
\bottomrule
\end{tabular}
}
\caption{Hyper-parameter settings for English and Chinese (CN) pre-trained model.}
\label{en_param}
\end{table}

\subsection{Baselines}

For English STS, we compare our methods to competitive contrastive methods including 1) \textbf{IS-BERT} maximizes agreement between sentence embeddings and token embeddings \cite{zhang2020unsupervised}, 2) \textbf{CT-BERT} uses different encoders to obtain augmented views \cite{janson2021semantic}, 3) \textbf{DeCLUTR} samples spans from the same document as positive pairs \cite{giorgi2021declutr}, 4) \textbf{ConSERT} designs four kinds of augmentation methods, i.e., adversarial attack, token shuffling, cutoff and dropout \cite{yan2021consert}, 5) \textbf{SimCSE} utilizes standard dropout as augmentation \cite{gao2021simcse} and 6) \textbf{ESimCSE} \cite{wu2021esimcse} introduces word repetition and momentum contrast mechanisms \cite{he2020momentum}, 7) \textbf{ArcCSE} \cite{zhang2022contrastive} modifies contrastive loss with angular margin and models entailment relation of triplet sentences, 8) \textbf{DiffCSE} \cite{chuang2022diffcse} adds an extra MLM-based word replacement detection task as an equivalent augmentation.

\begin{table*}[ht]
\centering
\resizebox{\textwidth}{!}{%
\begin{tabular}{lcccccccc}
\toprule
\textbf{Methods} & \textbf{STS12} & \textbf{STS13} & \textbf{STS14} & \textbf{STS15} & \textbf{STS16} & \textbf{STS-B} & \textbf{SICK-R} & \textbf{Avg.} \\
\hline\hline
\multicolumn{9}{c}{\textit{BERT Models}} \\
\hline
IS-BERT${_{base}}^\diamondsuit$ & 56.77 & 69.24 & 61.21 & 75.23 & 70.16 & 69.21 & 64.25 & 66.58 \\
CT-BERT${_{base}}^\diamondsuit$ & 61.63 & 76.80 & 68.47 & 77.50 & 76.48 & 74.31 & 69.19 & 72.05 \\
ConSERT${_{base}}^\spadesuit$ & 64.64 & 78.49 & 69.07 & 79.72 & 75.95 & 73.97 & 67.31 & 72.74 \\
SimCSE-BERT${_{base}}^\diamondsuit$ & 68.40 & 82.41 & 74.38 & 80.91 & 78.56 & 76.85 & 72.23 & 76.25 \\
ESimCSE-BERT${_{base}}^\clubsuit$ & 69.79 & 83.43 & 75.65 & 82.44 & 79.43 & 79.44 & 71.86 & 77.43 \\
ArcCSE-BERT${_{base}}^\spadesuit$ & 72.08 & 84.27 & 76.25 & 82.32 & 79.54 & 79.92 & 72.39 & 78.11 \\
DiffCSE-BERT${_{base}}^\heartsuit$ & \textbf{72.28} & 84.43 & 76.47 & 83.90 & 80.54 & \textbf{80.59} & 71.23 & 78.49 \\
\hline
\textbf{Our Methods} & & & & & & & & \\ 
\hline
Punctuation Insertion & 71.92 & 84.38 & \textbf{76.84} & \textbf{83.92} & 80.45 & 80.25 & 74.26 & \textbf{78.86} \\
Modal Verbs & 71.35 & 84.45 & 76.60 & 83.77 & \textbf{80.57} & 80.31 & \textbf{74.85} & 78.84 \\
Double Negation & 71.23 & \textbf{84.49} & 75.88 & 83.34 & 79.37 & 79.67 & 74.32 & 78.33 \\ 
Ensemble & 72.31$\pm0.38$ & 83.66$\pm0.64$ & 76.59$\pm0.38$ & 84.10$\pm0.20$ & 80.41$\pm0.42$ & 80.17$\pm0.52$ & 72.78$\pm0.49$ & 78.57$\pm0.43$ \\ 
\hline\hline 
\multicolumn{9}{c}{\textit{RoBERTa Models}}\\
\hline
DeCLUTR-RoBERTa${_{base}}^\diamondsuit$ & 52.41 & 75.19 & 65.52 & 77.12 & 78.63 & 72.41 & 68.62 & 69.99 \\
SimCSE-RoBERTa${_{base}}^\diamondsuit$ & 70.16 & 81.77 & 73.24 & 81.36 & 80.65 & 80.22 & 68.56 & 76.57 \\
ESimCSE-RoBERTa${_{base}}^\clubsuit$ & 69.90 & 82.50 & 74.68 & 83.19 & 80.30 & 80.99 & 70.54 & 77.44 \\
DiffCSE-RoBERTa${_{base}}^\heartsuit$ & 70.05 & 83.43 & 75.49 & 82.81 & 82.12 & 82.38 & 71.19 & 78.21 \\
\hline
\textbf{Our Methods} & & & & & & & & \\ 
\hline
Punctuation Insertion & 70.92 &	83.59 &	76.87 &	\textbf{83.73} &	\textbf{82.42} & \textbf{83.02} & 74.89 & 79.35 \\
Modal Verbs & \textbf{72.37} & \textbf{83.80} & 77.51 & 83.58 & 82.29 &	82.98 &	74.69 &	\textbf{79.60} \\
Double Negation & 71.07 & 83.56 & \textbf{77.60} & 83.38 & 81.59 & 81.82 &	\textbf{75.44} & 79.21 \\
Ensemble & 72.64$\pm0.62$  & 83.45$\pm0.38$  & 76.90$\pm0.17$  & 83.56$\pm0.21$  & 81.82$\pm0.38$  & 82.76$\pm0.31$  & 74.77$\pm0.6$  & 79.37$\pm0.07$ \\ 
\bottomrule
\end{tabular}
}
\caption{Performance comparison with existing competitive methods on STS tasks. The bold numbers top the competitions with the same pre-trained encoder and datasets. $\diamondsuit$: results from \cite{gao2021simcse}; $\clubsuit$: results from \cite{wu2021esimcse}; $\heartsuit$: results from \cite{zhang2020unsupervised}; $\spadesuit$: results from \cite{chuang2022diffcse}.}
\label{mr}
\end{table*}

\subsection{Training Details}
We use the same contrastive learning framework introduced by SimCSE, and train three models with Eq. \ref{eq4}. Each model uses one of the three augmentation methods to generate the positive sample and negation transformation to obtain the extra negative sample. Since not all sentences can be augmented with the rules as detailed in Table \ref{aug_per}, we pair the sentence with itself. 

Our implementation is based on \texttt{transformers} package. Following SimCSE, we use pre-trained checkpoints of English BERT\footnote{https://huggingface.co/bert-base-uncased} (uncased) \cite{devlin-etal-2019-bert} or RoBERTa\footnote{https://huggingface.co/roberta-base} \cite{liu2019roberta}. For the Chinese STS task, we use Chinese-RoBERTa-base\footnote{https://huggingface.co/hfl/chinese-roberta-wwm-ext} as our pre-trained model. Two MLP layers with batch normalization are added on top of the [CLS] representation in training but removed in evaluation. We train our model with Adam optimizer for one epoch and evaluate the model every 250 training steps on the development set and keep the best checkpoint for the final evaluation on test sets. The settings of important hyper-parameters are listed in Table \ref{en_param}.

\begin{table*}[ht]
\centering
\resizebox{\textwidth}{!}{%
\begin{tabular}{lcccccccc} 
\toprule
\textbf{Method}                                                 & \multicolumn{1}{c}{\textbf{MR}} & \multicolumn{1}{c}{\textbf{CR}} & \multicolumn{1}{c}{\textbf{SUBJ}} & \textbf{MPQA}  & \textbf{SST}   & \multicolumn{1}{c}{\textbf{TREC}} & \multicolumn{1}{c}{\textbf{MRPC}} & \multicolumn{1}{c}{\textbf{Avg.}}  \\ 
\hline
SimCSE-RoBERTa${_{base}}^{}\diamondsuit$ & 81.18 & 86.46 & 94.45 & 88.88 & 85.50 & 89.80 & 74.43 & 85.81 \\
ArcCSE-BERT${_{base}}^{}\spadesuit$      & 79.91 & 85.25 & \textbf{99.58} & \textbf{89.21} & 84.90 & 89.20 & 74.78 & 86.12 \\
DiffCSE-RoBERTa${_{base}}^{}\heartsuit$  & 82.82  & 88.61  & 94.32  & 87.71  & \textbf{88.63} & 90.40 & 76.81 & 87.04 \\
\hline
\textbf{Our Methods} &  &  &  &  &  &  &  &  \\
\hline
Punctuation Insertion & \textbf{83.59} & 87.79 & 93.81 & 88.10  & 87.81 & \textbf{91.60}   & \textbf{76.93}  & \textbf{87.09}  \\
Modal Verbs & 82.24 & \textbf{88.88} & 93.67 & 88.10 & 87.10 & 90.00 & 76.58 & 86.65  \\
Double Negation  & 81.73 & 87.26 & 93.61 & 88.12 & 87.26 & 89.00 & 77.28 & 86.32 \\
Ensemble  & 82.26$\pm0.8$ & 88.61$\pm0.29$ & 93.96$\pm0.59$ & 88.81$\pm0.07$ & 87.64$\pm0.82$ & 88.60$\pm0.60$ & 76.23$\pm0.93$ & 86.59$\pm0.17$         \\
\bottomrule
\end{tabular}
}
\caption{SentEval transfer tasks results of different models. $\diamondsuit$: results from \cite{gao2021simcse}; $\heartsuit$: results from \cite{zhang2020unsupervised}; $\spadesuit$: results from \cite{chuang2022diffcse}.}
\label{transfer}
\end{table*}

\section{Results and Analysis} \label{sec:results}

\subsection{Main Results} 
The experimental results including the three SDA methods are shown in Table \ref{mr}. Ensemble refers to the integration of three SDA methods. The implementation involves randomly selecting one augmentation method for each training sentence to generate positive samples. To eliminate the uncertainty introduced by the random seed, the experiment was conducted five times, and the ensemble results are presented as a range. In the case of BERT and RoBERTa, ensemble is represented as the average of three SDA methods. It should be noted that the results in the Table \ref{mr} have integrated standard negation as hard negative examples. 

Overall, experimental results clearly demonstrate that the proposed SDA methods lead to significant improvement in unsuperivsed sentence representation learning, which is evidenced by pushing state-of-the-art performance to a new bar of 79.60 with an overall improvement of 1.11\%. More specifically, when we look into those RoBERTa-based cases (Table \ref{mr}, the bottom half block), three proposed data augmentation methods (i.e., PI, MV, and DN) consistently surpass the previous state-of-the-art performance by a clear margin across all STS.

In BERT-based cases, all methods can slightly outperform to state-of-the-art methods across all datasets. We suspect the reason why our methods perform worse on BERT than on RoBERTa is caused by the introduction of next sentence prediction task (NSP) in pre-training BERT, which is not adopted in RoBERTa. The task predicts whether two sentences are semantically consecutive, making BERT more sensitive to the sentence-level semantic change as compared to RoBERTa. The learned representations from BERT are less likely to conform to the "semantic consistency" hypothesis. Previous works \cite{liu2019roberta,he2020deberta} also express similar opinions that NSP is not helpful for sentence representation. 

\begin{table*}[t]
\centering
\resizebox{\textwidth}{!}{%
\begin{tabular}{lcccccccc}
\toprule
\textbf{Methods} & \textbf{STS12} & \textbf{STS13} & \textbf{STS14} & \textbf{STS15} & \textbf{STS16} & \textbf{STS-B} & \textbf{SICK-R} & \textbf{Avg.} \\
\hline
SimCSE-RoBERTa${_{base}}$ & 70.16 & 81.77 & 73.24 & 81.36 & 80.65 & 80.22 & 68.56 & 76.57 \\
Modal Verbs (MV) & \textbf{72.37} & \textbf{83.80} & \textbf{77.51} & \textbf{83.58} & \textbf{82.29} & \textbf{82.98} & \textbf{74.69} & \textbf{79.60} \\
w/o Negation as Negative & \underline{71.80} & \underline{82.91} & \underline{76.19} & \underline{83.50} & 81.55 & \underline{82.37} & 68.78 & \underline{78.16} \\
w/o Positive Augmentation & 68.48 & 82.23 & 73.70 & 81.15 & \underline{81.56} & 81.26 & \underline{72.17} & 77.22 \\
\bottomrule
\end{tabular}
}
\caption{The effect of the positive augmentation and the negation transformation. The winners and runners-up are marked in bold font and underlined, respectively.}
\label{component}
\end{table*}

\begin{table}[t]
\centering
\begin{tabular}{lc}
\toprule
\textbf{Methods}    &\textbf{Chinese STS-B} \\ \hline
Roberta-base (last CLS)     &68.25  \\
SimCSE-RoBERTa-base         &71.10  \\
Punctuation Insertion      &71.98   \\
Modal Verbs                &\textbf{72.12} \\
Double Negation            &71.65   \\
\bottomrule
\end{tabular}
\caption{Performance comparison on Chinese STS task. All results come from our experiments, where SimCSE is the result we reproduced on Chinese dataset.}
\label{sts_zh}
\end{table}

It is interesting to find out that the performance of PI and MV are generally better than that of DN in most cases. These observations can be explained by our "semantic-expression trade-off'' hypothesis. Concretely, PI and MV introduce diversity in expression while retaining the semantic meaning of anchor sentences significantly. In contrast, the sophisticated correlation between linguistic expression (i.e., syntax, vocabulary) and the subtlety in semantics makes DN difficult to faithfully replicate the meaning of anchor sentences, deteriorating unsupervised learning performance. For example, "I'm hungry more'' is not semantically equivalent to "I'm not hungry no more'', and actually, they convey opposite meanings to some extent. Although we use rules based on syntactic analysis to limit the occurrence of this situation, it is difficult to completely eliminate it in practice. In the next section, we will conduct comprehensive ablation studies to elaborate on the importance of semantic consistency. 

\subsection{Transfer Tasks} 
Furthermore, to analyze the actual impact of semantic consistency on sentence representation, we evaluate the SentEval transfer tasks in Table \ref{mr}. The effect of the RoBERTa series models are better than that of BERT. So we selected the RoBERTa models to compare them with the previous state-of-the-art unsupervised contrastive learning models in transfer tasks using the SentEval toolkit\footnote{https://github.com/facebookresearch/SentEval}. For each task, SentEval trains the average embedding of the last two layers as the sentence embedding and evaluates the performance on the downstream task. To ensure a fair comparison, we did not introduce any additional training tasks, such as masked language modeling. As shown in Table \ref{transfer}, our SDA methods are able to match or surpass state-of-the-art approaches in certain tasks while falling behind other methods in some tasks. This suggests that different sentence representation learning methods excel at different downstream tasks, which may be related to their training paradigms. Nevertheless, our SDA method exhibits strong competitiveness in tasks other than the SUBJ. It's important to note that the results in Table \ref{mr} include the standard negation.

\subsection{Ablation Study}
To analyze the improvement made by the generated positive sample and the extra negative sample respectively, we compare four settings: using no discrete augmentation (SimCSE), using both positive augmentation (MV) and negation as negative, only using positive augmentation (w/o Negation as Negative) and only using negation as negative (w/o Positive Augmentation). We present the result of MV here in Table \ref{component} since its general good performance. We can observe that either using positive augmentation or using negation as negative achieves better performance compared to the method without discrete augmentation, and combining the two methods can make further improvements.

Notice that although using negation as negative performs seems worse on most STS series (STS12-16, STS-B) datasets than using positive augmentation, it obtains 3.39\% lift on SICK-R dataset. By measuring the average Jaccard similarity
\cite{jaccard1912distribution} between the test sentence pairs, we find that sentence pairs in SICK-R (57.24\%) are much more similar on text than the STS series of datasets (47.39\%), which supports our argument that textual similar negative sample can help alleviate feature suppression problem.

\begin{table}[t]
\centering
\begin{tabular}{llll}
\toprule
\textbf{Augmentations}  &       &       & \textbf{STS-B}          \\ \hline
None (Unsup-SimCSE)   &       &       & 82.45          \\ \hline
Cropping              & 10\%  & 20\%  & 30\%           \\
                      & 77.81 & 71.38 & 63.62          \\ \hline
Word Deletion         & 10\%  & 20\%  & 30\%           \\
                      & 75.89 & 72.20 & 68.24          \\ \hline
Synonym Replacement   &       &       & 77.45          \\
Mask 15\%             &       &       & 62.21          \\
Word Repetition       &       &       & 84.09          \\ \hline
Punctuation Insertion &       &       & 84.55          \\
Modal Verbs           &       &       & \textbf{84.99} \\
Double Negation       &       &       & 84.12          \\ 
\bottomrule
\end{tabular}
\caption{Performance comparison with discrete augmentation methods. Dropout is set to default in training as adopted in SimCSE and its variants.}
\label{tab:compare_vs_discrete}
\end{table}

\begin{figure*}[t]
\centering
\includegraphics[width=0.8\textwidth]{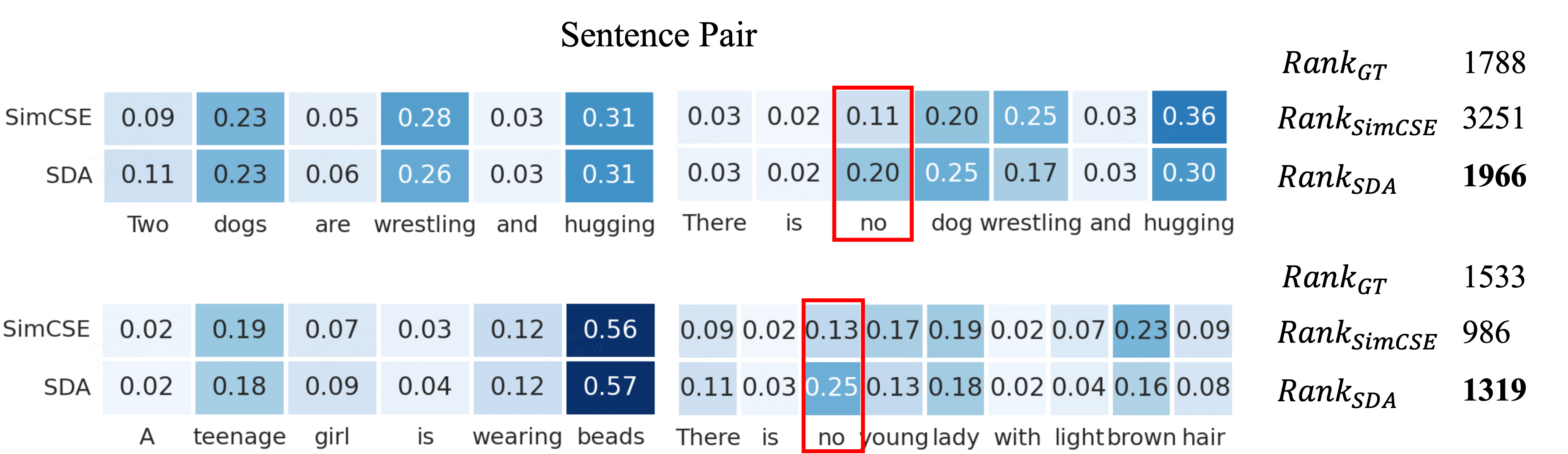}
\caption{Cases from SICK-R test set. The heatmap visualizes the spectrum of the weight of words in the sentence representation. We rank sentence pairs based on sorted similarity scores in ascending order. A better ranking should be closer to the ground truth (GT).}
\label{case}
\end{figure*}

\subsection{Results on Chinese Datasets}
Results on Chinese STS datasets are presented in Table \ref{sts_zh}. As compared to state-of-the-art method, PI, MV and DN gain an improvement of 0.78\%, 1.02\% and 0.55\% respectively, demonstrating that the proposed SDA methods can be seamlessly generalized to other languages like Chinese. This is consistent with our expectation since some linguistic phenomena are common in most of languages, e.g., minimal changes on punctuation or tone will not affect semantics too much.

\subsection{Random manipulation vs. rule-based augmentation}
We further compare different discrete augmentation methods such as cropping, word deletion, synonym replacement, masking, and word repetition as well as no discrete augmentation\footnote{When no discrete augmentation is applied, the setting falls back to SimCSE.} under the same contrastive learning framework. The results are shown in Table \ref{tab:compare_vs_discrete}. A comparison of the development set of STS-B shows that all the proposed augmentation methods outperform previous methods. Cropping, word deletion, synonym replacement, and masking \cite{wei2019eda,wu2020clear,meng2021coco} even perform worse than applying no discrete augmentation. We assume the worse performance is in part due to the randomness inserted by these manipulations, dramatically changing the sentence structure. Word repetition duplicates words in a sentence and rarely changes the semantics, thus it gets a competitive result, which again demonstrates the importance of keeping semantic consistency. 

We then try different manipulations where the randomness is reduced and thus semantic shift is smaller. More specifically, in Table \ref{pos}, we test three settings, no manipulation, inserting punctuation based on rules (i.e., PI), as well as randomly inserting punctuation (Random Insertion). Notice that we don't use negation as a negative sample in this experiment to control the variable. It turns out that applying augmentation according to the rules stands out among them, while Random Insertion performs worst. This result again verifies the importance of retaining semantic consistency since randomly inserted punctuation is more likely to disturb sentence semantics while rule-based PI generates syntax-correct sentences, boosting representation learning.

\begin{table}[t]
\centering
\begin{tabular}{lc}
\toprule
\textbf{Augmentations} & \textbf{avg. STS} \\ \hline
None (Unsup-SimCSE) & 76.57  \\
Random Insertion & 76.22    \\
Punctuation Insertion & 77.90   \\
\bottomrule
\end{tabular}
\caption{Performance comparison between random augmentation and rule-based augmentation.}
\label{pos}
\end{table}

\subsection{Case Study}

We calculate the weight of each word in the sentence representation by gathering the attention weights of the output token. We find that the learned representations of our method pay more attention to negative words. Since negative words have great influences on sentence semantics, SDA has largely excelled in those cases where negative words occurred, as exemplified in Fig. \ref{case}.

\begin{figure}[h]
\centering
\includegraphics[width=1\columnwidth]{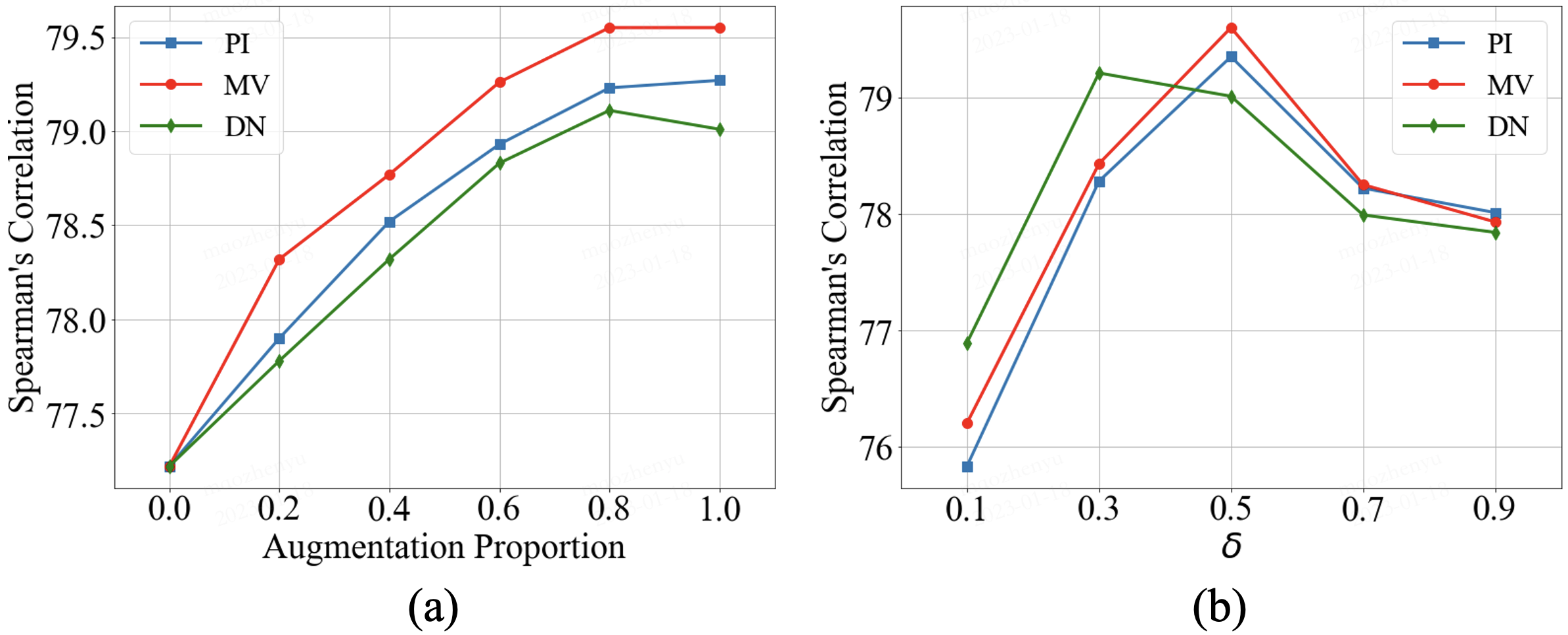}
\caption{Parameter sensitivity for (a) proportion of augmented sentence pairs in training and (b) margin $\delta$.}
\label{param}
\end{figure}

\subsection{Parameter Analysis}
We first compare the results with different augmentation proportions. When we set the augmentation proportion to $x\%$, only $x\%$ samples in the dataset are paired with augmentation, and $(1-x)\%$ samples are paired with themselves, which means $(1-x)\%$ samples use dropout as augmentation method. We can see from Fig. \ref{param} (a), the performance increases with a higher augmentation proportion. 

Then, we study how parameter $\delta$ influences our methods. By tuning $\delta$ rate from 0.1 to 0.9, we collect the results in Fig. \ref{param} (b). We find the best settings are 0.5 for both PI and MV, and 0.3 for DN. Since the augmented sentence constructed by DN is more similar to the negation sentence on text than the other two augmentation methods, thus a tighter restriction is needed to help differentiate the positive and negative.



\section{Conclusion}

In this work, we study the role of data augmentation in contrastive sentence representation learning and argue that the desiderata of reasonable data augmentation methods are to balance semantic consistency and expression diversity. Three simple yet effective discrete augmentation methods, i.e., punctuation insertion, modal verbs, and double negation are developed to substantiate the hypothesis. We further alleviate feature suppression with a negated sentence as a negative sample. 
Armed with well-executed discrete data augmentation, we achieve better results compared to more complicated state-of-the-art methods. Our work offers a promising direction of textual data augmentation and can be readily extended to multiple languages and beyond along this direction.

\section{Limitations}

Since our augmentation methods are implemented by rules, there could be some cases that the rules are not applicable or cause syntax mistakes. We have tried our best to alleviate such phenomenon and make the rule cover as many samples as possible as shown in Table \ref{aug_per}. We believe that it is absolutely feasible to use advanced LLMs like GPT4 to improve the quality of double negative generation, especially in cases where some simple methods do not work well. However, the cost of API calls or LLMs deployment cannot be ignored, especially for academia. Using LLMs to combine simple methods like ours to accelerate inference time or reduce resource dependencies is a valuable new direction.

In this work, we explore discrete text augmentation methods and propose three of them which achieve better results on STS tasks than previous studies. However, we believe that effective discrete augmentation methods can be abundant. Our purpose is not to find the best augmentation method, but to reveal the underlying rationales of a good augmentation method. Hence, we make our code available for replication and extension by the community.

\nocite{*}
\section{Bibliographical References}\label{sec:reference}

\bibliographystyle{lrec-coling2024-natbib}
\bibliography{refs}

\clearpage
\appendix

\section{Pseudo-Codes of Algorithms} \label{sec:algo}

Algorithm \ref{PI}, \ref{MV}, \ref{DN} are the pseudo-codes of the three proposed augmentation methods. The input sentence is first parsed by a spaCy sentence parser to obtain each token's part-of-speech (POS) and dependency tag. Then we apply the rules described in the algorithms on the parsed sentence to construct the augmented sentences.

\begin{algorithm}[!htbp]
	\renewcommand{\algorithmicrequire}{\textbf{Input:}}
	\renewcommand{\algorithmicensure}{\textbf{Output:}}
	\caption{Punctuation Insertion}
	\label{PI}
	\begin{algorithmic}[1]
        \REQUIRE one sentence $S$
	    \ENSURE the augmented sentence
		\IF{subordinate clause in $\mathcal{S}$} 
    		\STATE $pos \leftarrow$ Start position of subordinate clause
            \STATE return $S.\mathrm{Insert}(',', pos)$
		\ENDIF
		\IF{noun subject in $S$}
            \STATE $pos_e \leftarrow$ End position of the noun subject
            \STATE $S \leftarrow S.\mathrm{Insert}(',', pos_e)$ 
            \STATE return $S$
        \ENDIF
        \IF{$S$ end with punctuation $punc$}
            \STATE $S \leftarrow S.\mathrm{Replace}(punc, '!')$ 
        \ELSE
            \STATE $S \leftarrow S.\mathrm{Append}('!')$ 
        \ENDIF
        \STATE return $S$
	\end{algorithmic}
\end{algorithm}

\begin{algorithm}[!htbp]
	\renewcommand{\algorithmicrequire}{\textbf{Input:}}
	\renewcommand{\algorithmicensure}{\textbf{Output:}}
	\caption{Modal Verbs}
	\begin{algorithmic}[1]
        \REQUIRE one sentence $S$, set of modal verbs $\mathcal{V}_{mv}$
		\ENSURE the augmented sentence
		\STATE $mv \leftarrow \mathrm{Random.Choice}(\mathcal{V}_{mv})$
		\FOR{word $w$ in sentence}
            \IF{$w$ is be-verb}
                \STATE return $S.\mathrm{Replace}(w, mv + ' be')$
            \ENDIF
            \IF{$w.dep == ROOT$ and $w.pos == verb$}
                \STATE return $S.\mathrm{Replace}(w, mv + w.lemma)$
            \ENDIF 
        \ENDFOR 
        \STATE return $S$
	\end{algorithmic}  
\label{MV}
\end{algorithm}

\begin{algorithm}[!htbp]
	\renewcommand{\algorithmicrequire}{\textbf{Input:}}
	\renewcommand{\algorithmicensure}{\textbf{Output:}}
	\caption{Double Negation}
	\begin{algorithmic}[1]
        \REQUIRE one sentence $S$
		\ENSURE the augmented sentence
		\STATE $S' \leftarrow S$, $count \leftarrow 0$
		\FOR{word $w$ in $S$}
            \IF{$count == 2$}
		        \STATE return $S$
            \ENDIF
            \IF{$w$ is negative word}
        		\STATE $S \leftarrow S.\mathrm{Delete(w)}$
                \STATE $count++$
            \ENDIF
            \IF{$w.dep == aux$}
                \STATE $S \leftarrow S.\mathrm{Replace}(w, w + '\ not')$
                \STATE $count++$
            \ENDIF
            \IF{$w.dep == ROOT$ and $w$ is verb}
                \STATE $S \leftarrow S.\mathrm{Replace}(w, 'do\ not\ ' + w)$
                \STATE $count++$
            \ENDIF
        \ENDFOR
        \IF{$count == 1$}
		    \STATE return $'Not\ that\ ' + S$
	    \ELSE
	        \STATE return $S'$
        \ENDIF
	\end{algorithmic}  
\label{DN}
\end{algorithm}

\end{document}